\documentclass[a4paper, 10pt, conference]{ieeeconf}      
\usepackage{FG2021}

\FGfinalcopy 

\IEEEoverridecommandlockouts                              
\usepackage{graphics} 
\usepackage{epsfig} 
\usepackage{mathptmx} 
\usepackage{times} 
\usepackage{amsmath} 
\usepackage{amssymb}  
\usepackage{caption}
\usepackage[accsupp]{axessibility}

\def\FGPaperID{63} 

\usepackage{xspace}
\newcommand{\SignLookup}{\emph{Sign-Lookup}}
\newcommand{\etal}{\emph{et al.}\xspace}

\newcommand{\ra}[1]{\renewcommand{\arraystretch}{#1}}
\def\B{\fontseries{b}\selectfont}

\title{\LARGE \bf
Looking for the Signs: Identifying Isolated Sign Instances in Continuous Video Footage
}

\author{\parbox{16cm}{\centering
    {\large Tao Jiang, Necati Cihan Camg\"{o}z, Richard Bowden}\\    {\normalsize
    Centre for Vision, Speech and Signal Processing, \\
University of Surrey, Guildford, UK}}
}

\begin{document}

\ifFGfinal
\thispagestyle{empty}
\pagestyle{empty}
\else
\author{Anonymous FG2021 submission\\ Paper ID \FGPaperID \\}
\pagestyle{plain}
\fi
\maketitle

\begin{abstract}
   In this paper, we focus on the task of one-shot sign spotting, i.e. given an example of an isolated sign (\textit{query}), we want to identify whether/where this sign appears in a continuous, co-articulated sign language video (\textit{target}). To achieve this goal, we propose a transformer-based network, called \SignLookup. We employ 3D Convolutional Neural Networks (CNNs) to extract spatio-temporal representations from video clips. To solve the temporal scale discrepancies between the query and the target videos, we construct multiple queries from a single video clip using different frame-level strides. Self-attention is applied across these query clips to simulate a continuous scale space. We also utilize another self-attention module on the target video to learn the contextual within the sequence. Finally mutual-attention is used to match the temporal scales to localize the query within the target sequence. Extensive experiments demonstrate that the proposed approach can not only reliably identify isolated signs in continuous videos, regardless of the signers' appearance, but can also generalize to different sign languages. By taking advantage of the attention mechanism and the adaptive features, our model achieves state-of-the-art performance on the sign spotting task with accuracy as high as 96\% on challenging benchmark datasets and significantly outperforming other approaches.
   \end{abstract}

\section{Introduction}
\label{sect:intro}
Sign Languages are the native languages among Deaf communities. They are languages in their own right, distinct from spoken language but as linguistically complex as any spoken language. Each country has its own sign language often with regional variations. They incorporate manual (including handshape and motion) and non-manual (facial expression and body posture) channels or articulators which are combined via grammatical constructs that use both direction and space to convey meaning. As such, the grammar and word ordering of sign is very different to spoken language. 

Although sign language dictionaries often give isolated forms of a sign, without any co-articulation or contextual modification,  a continuous sign sequence is not just a concatenation of those isolated signs in spoken word order. Often, the first stage of linguistic annotation is to annotate this sequence of sign glosses contained in the video. Thus, the ability to identify examples of signs within a continuous stream would be a valuable tool, both for linguistic analysis and computational sign language research.

In this work, we focus on the sign lookup problem, that is, looking for an isolated sign in a continuous, co-articulated sign language video. This process, often called \emph{Sign Spotting}, has a wide range of applications e.g. searching in sign videos by keyword to allow a content-aware search; creating or augmenting isolated sign dictionaries from other unlabeled footage; semi-automatic annotation (glossing) of sign language data for linguistic study; or used in assessment systems to automatically check whether an L2 Language learner is signing correctly.

One possible way to achieve this goal is to extract features from the \emph{query} and the \emph{target} videos, and then compute the feature similarity. However, directly applying this method will result in poor performance. This is because the \emph{query} and the \emph{target} videos may have severe spacial and temporal domains differences: (1) Signers in the \emph{query} and \emph{target} videos may be different people and/or have different appearance. (2) Different signers have their own signing habits or style, using slightly different motions even though they perform the same sign. (3) The sign speed in the \emph{query} and the \emph{target} videos may also be different. (4) The direction of the sign may be modified from the citation form of the sign (5) For one word, there might be multiple corresponding signing forms.

To tackle these challenges, we propose a Transformer-based deep learning model for sign lookup. We first utilize a 3D CNN to learn the spatio-temporal features from sign video clips. We pre-train this module  for sign recognition using both the isolated signs (\emph{query}) and continuous sign phrases (\emph{target}). Once trained, the last classification layer is removed, and the 3D-CNN model is used as a feature extractor for the transformer model. To solve the temporal scale difference, we adopt different strategies while extracting features for the \emph{query} and the \emph{target} videos. For the isolated \emph{query} sign, we extract 4 features with different temporal scales. Neighbouring frames in a window are separated from each other by a temporal offset of 1, 2, 4 or 8 frames. In this way, the extracted \emph{query} features cover 4 different temporal scales, which we term adaptive features. For the continuous \emph{target} video, a sliding window approach is applied to extract features throughout the whole video. Then self-attention is applied on the adaptive \emph{query} features and the sliding \emph{target} features. For the \emph{query}, this simulates a continuous temporal scale space by learning a linear combination of discrete temporal scales. For the \emph{target}, it learns the contextual relationship within the continuous sign video. In addition, mutual-attention learns to match multi-time scaled queries within the target sequence. This decreases the temporal scale difference between the \emph{query} and the \emph{target}. Finally, a multi-layer perceptron is use for classification. 

The contributions of this paper can be listed as: (1) we propose a novel Transformer-based neural network for sign lookup and demonstrate its application to sign spotting. (2) we extract adaptive features for a bank of isolated videos and apply self-attention, which simulates a continuous temporal scale space. (3) we apply mutual-attention on the adaptive \emph{query} and \emph{target} features, which closes the temporal domain difference gap between the \emph{query} and the \emph{target}. (4) We demonstrate generalization to two different sign languages and experiments demonstrate state-of-the-art performance.

The rest of the paper is structured as follows: In Section~\ref{sect:litrew} we cover the related work on sign language recognition, sign spotting and sign language datasets. In Section~\ref{sect:method} we describe the \SignLookup~architecture and its components in detail. We describe the training process and implementation details in Section~\ref{sect:expdetails}. In Section~\ref{sect:quant} and Section~\ref{sect:qual} we perform qualitative and quantitative experiments, respectively, to demonstrate the effect of \SignLookup~on sign language learning. We apply our model to the downstream task of sign spotting, achieving state-of-the-art performance. Finally, we conclude the paper in Section~\ref{sect:conc} by discussing our findings and the future work. 

\section{Related work}
\label{sect:litrew}
Our work is related to several topics in the literature including the curation of sign language datasets as well as sign language recognition and sign spotting. We will discuss each in turn.

\noindent \textbf{Sign language datasets:} There is no universal sign language. Each country has its own distinct language and may have regional dialects that have considerable difference even within a single country. However, some languages are better represented by public datasets than others. For example, there are many public datasets that have been released for American Sign Language (ASL) \cite{athitsos2008american,joze2018ms,li2020word,wilbur2006purdue}, German Sign Language (DGS)\cite{koller2015continuous,von2008significance}, Chinese Sign Language (CSL) \cite{chai2014devisign,huang2018video}, Finnish Sign Language (FSL) \cite{viitaniemi2014s} and British Sign Language (BSL) \cite{albanie2020bsl,schembri2013building} to name but a few. For BSL, Albanie \etal \cite{albanie2020bsl} use mouthing cues to collect a large-scale dataset of co-articulated BSL signs. However, this data is not currently publicly available. A large BSL dataset that is available and partially annotated is the BSLCORPUS captured by Schembri \etal \cite{schembri2013building}. The BSLCORPUS consists of continuous sign videos with fine-grained linguistic annotation. Felon \etal \cite{fenlon2014bsl, fenlon2015building} created an isolated BSL dictionary, for which the signs are consistent with the BSLCORPUS. As this provides a paired corpus of isolated signs and continuous video containing those signs, it is ideal for our application of sign spotting.

\noindent \textbf{Sign language recognition:} Automatic sign recognition has drawn a lot of attention from the computer vision research community in the last 30 years. Early work on sign language recognition \cite{farhadi2007transfer,fillbrandt2003extraction,tamura1988recognition,kadir2004minimal} focused on designing hand-crafted features that describe hand shape and motion. But to model the temporal variation of sign, approaches often mimic those developed for speech recognition. For example, Hidden Markov Models (HMMs) have been  used widely \cite{kadir2004minimal, starner1995visual, forster2013modality, von2008recent}. 

However, with the appearance of deep learning, neural networks have become popular. CNNs have demonstrated significant improvement in manual \cite{koller2016deep1, koller2016deep2} and non-manual \cite{koller2014read, koller2015deep} feature extraction. RNNs have shown superior performance to HMMs at learning dynamic temporal dependencies \cite{mitra2007gesture, gupta2016online, ye2018recognizing}, but one problem for RNNs is they are prone to vanishing gradients. To overcome this problem, LSTM \cite{yao2015describing, guo2018hierarchical,pan2016jointly} and GRU \cite{camgoz2018neural} use gating to regulate the information flow. However, the information flow in an LSTM is one way: the current state can only inherent from its previous states; the following states have no effect on the current state. To overcome this, bidirectional LSTM (BLSTM) \cite{camgoz2017subunets, huang2018video, zhou2020spatial, cui2017recurrent} combine one forward LSTM and one backward LSTM such that the current state has full knowledge of the entire sequence. Just as 2D CNNs are ideal for image data, 3D CNNs \cite{bilge2019zero, camgoz2016using, huang2015sign, ye2018recognizing} have been applied to video sequences. When modelling spatio-temporal data, the depth dimension in 3D CNNs represents the temporal dimension of the sequence. I3D \cite{carreira2017quo} based methods \cite{albanie2020bsl, joze2018ms, li2020transferring} have shown significant performance improvements in sign language learning problems. In this paper, we use I3D to extract the spatio-temporal features.

\noindent \textbf{Sign language spotting:} Viitaniemi \etal \cite{viitaniemi2014s} crafted a sign descriptor based on a $5 \times 5$ skin distribution histogram, and used dynamic time warping for temporal alignment. Ong \etal \cite{ong2014sign} modeled the spatio-temporal signatures of a
sign using Sequential Interval Patterns, and then organized them in a hierarchical tree structure for classification. According to the number of the instances available in the dictionary, sign spotting can be categorised as: zero-shot, one-shot or few-shot. For zero-shot learning, Bilge \etal \cite{bilge2019zero} constructed spatio-temporal
representations by using 3D CNNs + LSTMs, and then leveraged descriptive text embeddings to discover the unseen sign. Besides one-shot examples, work in \cite{pfister2014domain} utilised a weakly supervised reservoir containing multiple instances of a gesture to significantly improve the performance of a gesture classifier. Momeni \etal \cite{momeni2020watch} employed multiple-instance contrastive learning to solve the few-shot problem combined with other supervisory signals: sparse annotations and subtitles. In our case, where we have only a single \emph{query} example of a sign, it is a one-shot learning problem.

\begin{figure*}%
    \centering
    \includegraphics[width=0.85\textwidth]{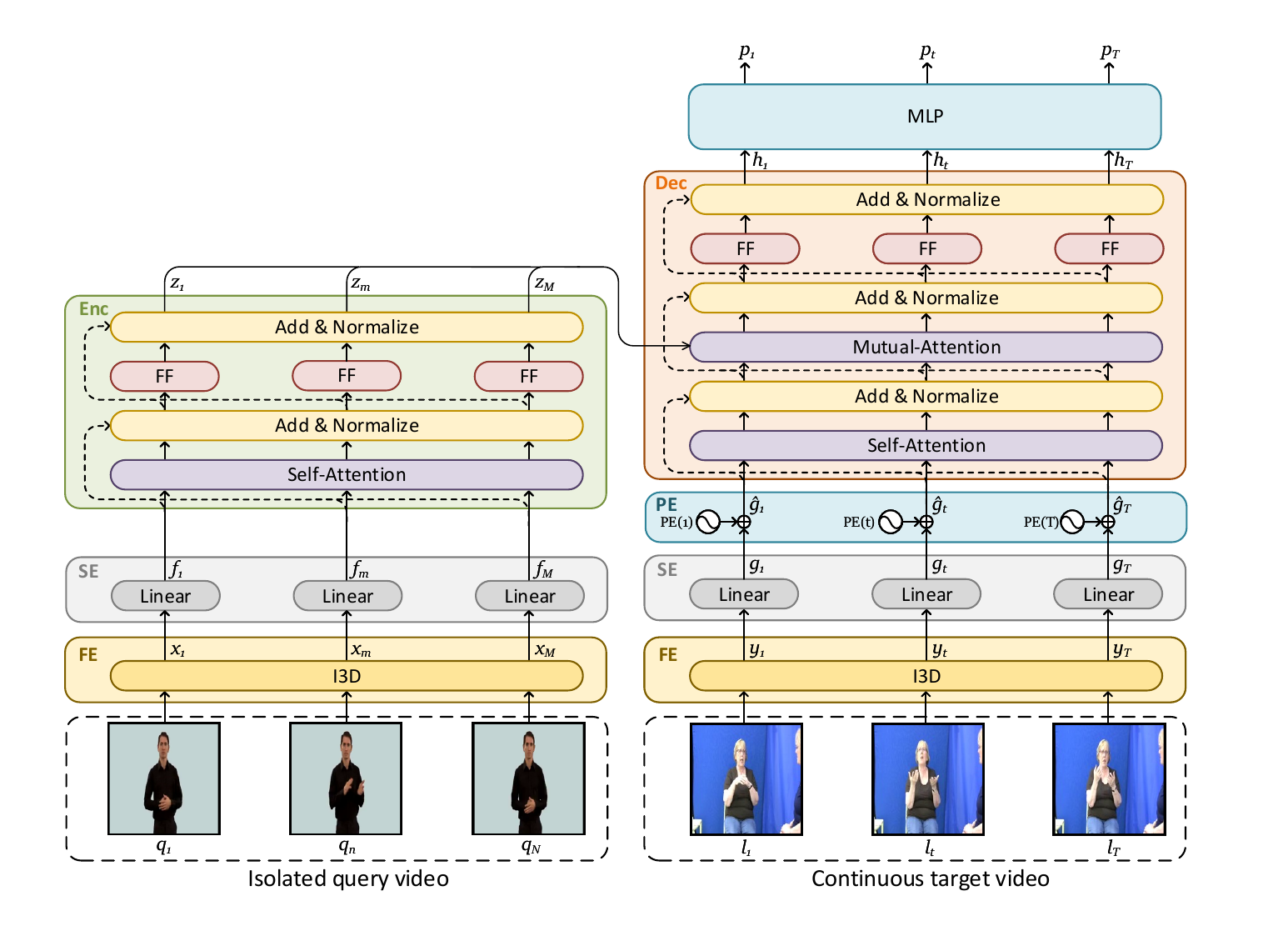}
    \captionsetup{justification=centering}
    \caption{ A detailed overview of a single layered of Sign-Lookup.\\
        (FE: Feature Extracting, SE: Spatial Embedding, PE: Positional Encoding, FF: Feed Forward, Enc: Encoder, Dec: Decoder)}
    \label{fig:overview}
\end{figure*}

\section{Sign-Lookup Architecture}
\label{sect:method}
In this section, we introduce the architecture of \SignLookup, a deep learning network capable of identifying an isolated sign \emph{query} within a continuous \emph{target} sign video. Based on the extracted spatio-temporal features, our approach considers not only the temporal dependencies in a \emph{target} video, but also the mutual dependencies between the \emph{query} and \emph{target} video.

The architecture of \SignLookup~is shown in Figure~\ref{fig:overview}. \SignLookup~consists of three tiers of neural network. Firstly, a 3D Convolutional Neural Network (CNN) takes a sequence of images as input and extracts spatio-temporal features for both the \emph{query} and \emph{target} videos. Secondly, self-attention is applied to the \emph{query} (adaptive) and the \emph{target} features to exploit the temporal scale in the \emph{query} video and the context in the \emph{target} video. Mutual attention is employed to learn the dependencies between them. Finally, for classification, a multi-layer perceptron (MLP) is applied to make the final frame-level predictions. In the rest of this section, we give more details on each tier of \SignLookup.

\subsection{Feature Extraction: 3D Convolutional Neural Networks}
\SignLookup~uses 3D CNNs to learn the spatio-temporal feature representations. Due to its successful application in human action recognition \cite{farha2019ms} and sign recognition \cite{albanie2020bsl,joze2018ms}, we adopt the I3D architecture \cite{carreira2017quo} to extract spatio-temporal features from the videos. Theoretically, the I3D model can take any length sequence of consecutive frames as input. However, according to \cite{buehler2009learning,pfister2013large,viitaniemi2014s}, co-articulated signs roughly last for 13 frames, so, in our experiments, we set the window size of I3D to 16 frames. For a \emph{query} video, we extract $M$ adaptive features. For the $m$-th level feature, based on the middle frame of the video, we take a frame every $2^{(m-1)}$ frames bilaterally up to 16 frames. For the \emph{target} video, we calculate a feature for each frame via a sliding window.

\subsection{Learning Dependencies: Transformer}
Sign spotting can be summarized as calculating the similarity  between the features of the \emph{query} and \emph{target} sequence: the more similar they are, the higher the chance the \emph{query} is present in the \emph{target}. However, we cannot just apply a dot product to the extracted features to calculate similarity, because: (1) The length of a signs vary and (2) The temporal scale is different between the \emph{query} and the \emph{target}.

To solve these problems, we propose a Transformer-based network. Assuming that we have a \emph{query} video $Q=(q_1, q_2, \cdots, q_N)$ with $N$ frames and a \emph{target} video $L=(l_1, l_2, \cdots, l_T)$ with $T$ frames, we extract $M$ adaptive features $X = (x_1, x_2, \cdots, x_M)$ for the \emph{query} video and $T$ features $Y = (y_1, y_2, \cdots, y_T)$ for the \emph{target} video via I3D (Fig1:FE). As can be seen in Figure~\ref{fig:overview} the features $x_m$ and $y_t$ are then projected into an embedding space through linear layers:
\begin{equation}
   f_m = W^f \cdot x_m + b^f, ~~~~g_t = W^g \cdot y_t + b^g
\end{equation}

To make the network aware of the frame ordering in the \emph{target} sequence, we also use positional encoding \cite{vaswani2017attention} on the \emph{target} embeddings:
\begin{equation}
   \hat{g}_t = g_t + \mathrm{PositionalEncoding}(t)
\end{equation}
where PositionalEncoding is a predefined sinusoidal function conditioned on the sequence position $t$. 

Next, the adaptive embedding vectors $F = (f_1, f_2, \cdots, f_M)$ are fed into an encoder, where they are self-attended to learn a continuous temporal scale:
\begin{equation}
   z_m = \mathrm{Enc}(f_m | f_{1:M})
\end{equation}

where $z_m$ denotes the contribution of the $m$-th basis $f_m$ to the temporal scale space, which is generated by the encoder on the $m$-th time scale level, given the adaptive embeddings of the query video $f_{1:M}$. 

The encoder consists of a stack of $n$ identical layers. Each layer contains two sub-layers: Multi-Head Attention (MHA) and Feed-Forward (FF) sub-layer. MHA produces a weighted contextual representation, performing multiple projections of scaled dot-production attention. Scaled dot-production attention is a linear combination of values $V$, weighted by the relevant queries $Q$, keys $K$ and dimensionality $d_k$:
\begin{equation}
   \mathrm{Attention}(Q,K,V)=\mathrm{softmax}(\frac{QK^T}{\sqrt{d_k}})V
\end{equation}
The second Feed Forward sub-layer consists of two linear layers with ReLU activation in-between (Fig 1:FF). 

The positional encoded vectors $\hat G = (\hat g_1, \hat g_2, \cdots, \hat g_T)$ are fed into a decoder (Fig1:Dec). The decoder contains a self-attention module and a mutual-attention module. Self-attention learns the contextual relationships within the target sequence. Then mutual-attention learns the temporal scale relationship between the \emph{query} and  \emph{target}. The key and value of mutual attention are the \emph{query} outputs from the encoder, while the query of mutual attention is the target representation after self-attention. We formulate the decoding process as:
\begin{equation}
   h_t = \mathrm{Dec}(\hat g_t | \hat g_{1:T}, z_{1:M} )
\end{equation}
where $h_t$ denotes the projection of the $t$-th \emph{target} feature in the temporal scale space of the \emph{query}.

\subsection{Classification: Multi-layer Perceptron}

From these learned embeddings, we want a projection to the final classes which are dependant upon the final  application. In the case of sign spotting, we have a binary decision indicating whether the \emph{query} sign appears in the \emph{target} video or not. To that end, we apply a multi-layer perceptron to the embeddding. The MLP consists of 5 fully connected layers with Leaky Relu activation functions between them. The first linear layer has a residual connection on the 2048-dimensional embedding. The dimensionality of the embedding is then gradually reduced to 1024, 512, 256 and finally 2 for binary classification:
\begin{equation}
   p_t = \mathrm{MLP}(h_t)
\end{equation}
where $p_t$ denotes the probability of the query appearing in the $t$-th frame of the \emph{target} video.

\section{Training and Implementation Details}
\label{sect:expdetails}
In this section, we discuss the implementation details necessary for training \SignLookup. First, we describe the sign video dataset used and then the pre-processing stages. As we use I3D for feature extraction, we next explain how to pre-train I3D before finally giving details of the transformer model parameters and its training process for sign spotting.

\subsection{Sign Video Dataset}
Our model makes use of two data streams: isolated sign videos which form the \emph{query} and continuous \emph{target} sign videos in which we seek to find the \emph{query} sign. For the isolated sign videos, we use the publicly available BSL SignBank Dictionary introduced by \cite{fenlon2014bsl, fenlon2015building}. SignBank consists of 3688 lexical signs from BSL (i.e. signs that are highly conventionalised in both form and meaning across the sign language community). Each video in SignBank contains one prototypical canonical and isolated sign. However, most of the videos in SignBank are isolated examples, the signer remains still at both the start and the end lacking any form of co-articulation. As we do not want to include these useless frames during training we use OpenPose \cite{cao2019openpose} to detect the signer's hand movement and trim the beginning and the end of the video where the signer is silent. Then, the trimmed videos are used for both I3D pre-training and \SignLookup~training. For the continuous videos, we use the ongoing BSL linguistic corpus BSLCORPUS introduced by \cite{schembri2013building}. The BSLCORPUS provides continuous BSL footage filmed by 249 Deaf participants from eight cities across the United Kingdom. It has partial human annotation for linguistic purposes. For sign spotting, we employ the gloss\footnote {A gloss is a written or typed approximation of a sign, typically using English words as “labels” for each sign and normally relating to a set lexicon.} annotation, which annotates approximate 72k signs with sign categories and temporal boundaries. For the annotated sign categories, there is correspondence between the SignBank lexicon and the gloss information annotated in the BSLCORPUS, This means we can use the isolated examples in SignBank to match against continuous examples in the BSLCORPUS. On this basis, we are only interested in the sign categories that are common to both datasets. Note that one English word might have different signing forms. Fortunately, SignBank and the BSLCORPUS distinguish these different forms consistently using different glosses. There are 2026 common sign categories for both datasets which we use in our tests. For pre-processing, we resize all videos in both datasets to 256 by 256 pixels with a fixed frame-rate of 25fps and then split the BSLCORPUS with a ratio of 4:1:1 for training, validation and test. There are 204 signers in training subset, 63 signers in validation subset and 64 signers in test subset. 39 signers  overlap between training and validation subsets.
 
To demonstrate the generalization of \SignLookup~across different sign languages, we also apply our model to DGS (German Sign Language). To do so we use the popular German Sign Language dataset: PHOENIX14T \cite{camgoz2018neural}, which covers unconstrained DGS sign from 9 different signers with a vocabulary of 1066 different signs and translations into spoken German with a vocabulary of 2887 different words. It composes a parallel corpus including sign languages videos, sign-gloss annotation and German translation, which are all segmented into parallel sentences. For our purpose, we are only interested in the sign-gloss annotations.
 
\subsection{I3D Pre-training}
I3D was originally designed for action recognition so before using I3D as a feature extractor for sign, we need to pre-train the I3D model on sign.  Videos in SignBank are isolated and can be used for training directly. However, videos in the BSLCORPUS are continuous and cannot be used directly. To adapt them for I3D training, we segment the continuous videos into individual sign clips according to the annotated timeline for the corpus. We only do this for the training subset of the BSLCORPUS. Then, both SignBank training and the isolated BSLCORPUS videos are used for I3D pre-training. After pre-training, the last linear layer of I3D is removed and the rest of the parameters are frozen for feature extraction. In our case, the extracted feature vector has 1024 dimensions.

\subsection{Sign-Lookup Training}
When constructing the training dataset for \SignLookup, we take four clips of 16 frames with 1, 2, 4, 8 steps from SignBank videos and a clip of 64 frames as a sliding window from BSLCORPUS videos with a stride of 32 frames. To ensure the data is balanced, we ensure that for every positive sample added to training, we add a negative sample (i.e. a clip from the BSLCORPUS that does not contain the sign). The ground truth is \emph{true} if the frame in the BSLCORPUS clip is annotated as the same sign as the SignBank example, \emph{false} otherwise. 

Once pre-trained, the I3D model is employed for feature extraction. In a SignBank video, we have four adaptive feature vectors. For the 64 frames in a BSLCORPUS clip, 64 feature vectors are extracted by applying I3D as a sliding window. We use Binary Cross-Entropy for classification which is minimized by Stochastic Gradient Descent (SGD) with mini-batches of size 8, and an initial learning rate of $10^{-2}$ with a plateau scheduler. This has a patience of 20 steps and a decrease factor of 0.1. All parameters in the network are initialized by xavier initialization \cite{glorot2010understanding}. Training was done on machines with 128GB RAM and a Nvidia Titan X GPU.

\subsection{Evaluation Metrics}
Following the works of \cite{farha2019ms,lea2017temporal}, we evaluate all methods using the metrics of frame-wise accuracy (Acc) and the $F1$ score at overlapping thresholds $k$ denoted by $F1@k$. The frame-wise accuracy is the ratio of correctly predicted frames to total frames, which has been widely used in action detection and segmentation. It penalizes the prediction disorder but is insensitive to over-segmentation errors. To overcome this, we use $F1@k$ score proposed by \cite{lea2017temporal}. It also allows minor temporal shifts, which can be caused by annotator variability. The $F1$ score is the harmonic mean of precision and recall: $F1 = 2 \frac{prec*recall}{prec+recall}$. The intersection over union (IoU) overlap between the prediction and the ground truth is compared to the threshold $k$ to decide a true or false positive, which is used for precision and recall calculation.

\section{Quantitative Experiments}
\label{sect:quant}
In this section, we evaluate our model on different tasks and report quantitative results. We first apply our model to sign spotting with different ablation configurations to demonstrate the importance of each component in our network. We then compare our model to previous work on sign-spotting task. Finally, we evaluate our model on other sign language datasets to show the generalization ability of \SignLookup~amongst different sign languages.

\subsection{Sign Spotting}                                                                         
We evaluate our model on sign spotting and investigate the effect of (1) the number of heads in an attention layer, (2) the number of encoder/decoder layers, (3) different batch sizes, (4) the adaptive features, (5) different levels of dropout. In the following experiments, we use 4 attention heads, 8 encoder/decoder layers. We train our models using a batch size of 4 and without any dropout unless specified otherwise.

\subsubsection*{Number of Attention Heads}
Compared to classical dot product attention, multi-head attention allows the model to learn and attend to multiple representation subspaces. In this experiment, we evaluate the effect of the number of heads in an attention layer. We train four networks with different numbers of attention heads (1, 2, 4, 8). 

\begin{table}[!h]\centering
\ra{1.05}
\caption{Comparing different numbers of attention heads.}\label{tbl:sspotting:nheads}
\resizebox{0.62\columnwidth}{!}{
{\normalsize
\setlength{\arrayrulewidth}{.06em}
\begin{tabular}{@{}c|c|ccc}
\# Heads&   Acc & \multicolumn{3} {c}{F1@\{25,50,75\}}   \\ \hline
1 & 91.4 & 79.4   & 69.8  & 50.4    \\
2 & 91.7 & 80.0   & 70.7 & 51.6     \\
4 & \B 91.8 & 80.3   & \B 70.8  & \B 51.9   \\
8 & 91.7 & \B  80.4      & 70.7    & 51.8       
\end{tabular}
}
}
\end{table}

As can be seen in Table~\ref{tbl:sspotting:nheads}, increasing the number of heads initially improves the performance. However, after 4 heads, the performance starts to decrease. We believe this is because the exploration in representation sub-spaces becomes saturated and introducing more heads after that point does not bring any performance enhancement but rather encourages over-fitting. We will use 4 attention heads in the remainder of our experiments.

\subsubsection*{Number of Encoder/Decoder Layers}
An encoder can consist of a stack of $n$ identical encoder layers. The lower layers contain more local features while the higher layers contain more global features. Theoretically, the more layers in the encoder, the more context is exploited in the embedding. However, too many can cause other problems like over-fitting and inefficiency. In this experiment, we train our model with different encoder layers to identify the trade-off and the optimal configuration. Encoders with 1, 2, 4, 8 layers are trained respectively. All results are shown in Table~\ref{tbl:encoderlayers}.

Similar to attention heads, increasing the number of layers initially improves performance. However, after 4 layers, the performance starts to decrease. We believe this is because more layers give us more global features at the beginning, which provides more information to the network for classification. After a certain point, global feature learning becomes saturated and introducing more heads can only lead to over-fitting.  We will use four encoder/decoder layers in the remainder of our experiments.

\begin{table}\centering
\ra{1.05}
\caption{Comparing different number of layers.}\label{tbl:encoderlayers}
\resizebox{0.65\columnwidth}{!}{
{\normalsize
\setlength{\arrayrulewidth}{.06em}
\begin{tabular}{@{}c|c|ccc}
\# Layers    & Acc & \multicolumn{3} {c}{F1@\{25,50,75\}}   \\ \hline
1 & 91.8 & 77.9   & 68.4  & 49.4    \\
2 & 91.8& 80.1   & 70.6  & 51.6     \\
4 & \B 91.8& \B 81.1   & \B 71.8  & \B 52.6     \\
8 & 91.8 & 80.3   &  70.8  &  51.9     
\end{tabular}
}
}
\end{table}

\subsubsection*{Batch Size}
Larger batch sizes tend to produce smoother gradients at the price of decreased convergence rate. On the contrary, smaller batch sizes can enhance the convergence rate but introduces more noise to training. However, according to the recent studies on the information theory behind deep learning \cite{tishby2015deep,shwartz2017opening}, the noise introduced by smaller batch size can help the network to represent the data more efficiently. In this experiment, we train our network with different batch sizes subject to the GPU memory limits. All the results are shown in Table~\ref{tbl:batchsizes}.

From the table, we can see that the largest batch size does not achieve the best performance, neither does the smallest. Too large a batch size slows convergence which tends to converge to a local optima. While too small, introduces too much noise, which can diminish performance. In the remainder of our experiments, we will use a batch size of 8.

\begin{table}\centering
\ra{1.05}
\caption{Comparing different batch sizes.}\label{tbl:batchsizes}
\resizebox{0.65\columnwidth}{!}{
{\normalsize
\setlength{\arrayrulewidth}{.06em}
\begin{tabular}{@{}c|c|ccc}
Batch Sizes  & Acc & \multicolumn{3} {c}{F1@\{25,50,75\}}     \\ \hline
1 & 91.1 & 78.2   & 68.4  & 49.4     \\
2 & 91.8 & 81.0   & 71.7  & 52.5    \\
4 & 91.8 & 81.1   &  71.8  &  52.6     \\
8 &\B 92.0 & 80.8    & \B 71.9    & \B 52.8        \\
16 & 91.8 &  \B 82.0      & 69.9    & 50.9       
\end{tabular}
}
}
\end{table}

\subsubsection*{Adaptive features}
In our model, we use adaptive features for the query video because of the temporal scale difference between the \emph{query} and the \emph{target} videos. To show the advantage of adaptive features, we compare to using 
only one feature for the \emph{query} video:  a single feature extracted from its middle 16 frames. With a single feature in the \emph{query}, we need to change our network structure. Firstly removing the encoder attention as there are no other features to attend to and secondly to replace the mutual-attention with a dot and element-wise product. The results are shown in Table~\ref{tbl:diffquery}.

One obvious observation from this experiment is that our method with adaptive features outperforms the method with only one feature. This is because using only one feature assumes the temporal scale is the same in the \emph{query} and \emph{target} videos, which is not true in this scenario. Our method takes advantage of adaptive features to reduce the temporal scale difference and achieve better results. 

\begin{table}\centering
\ra{1.05}
\caption{Comparing using single query feature and adaptive query features.}\label{tbl:diffquery}
\resizebox{0.78\columnwidth}{!}{
{\normalsize
\setlength{\arrayrulewidth}{.06em}
\begin{tabular}{@{}l|c|ccc}
Methods   & Acc & \multicolumn{3} {c}{F1@\{25,50,75\}}   \\ \hline

One feature & 91.1 & 66.8   & 58.1  & 42.1    \\
Adaptive features & \B 92.1 &  \B 84.9      & \B 75.7    & \B 56.0  \\
\end{tabular}
}
}
\end{table}

\subsubsection*{Dropout}
In this experiment, we train our model with dropout ranging from 0.0 to 0.5. All results are shown in Table~\ref{tbl:dropout}. As can be expected, the method with zero dropout produces poor results with 0.3 providing optimal results.

\begin{table}\centering
\ra{1.05}
\caption{Evaluating effects of different dropout rates.}\label{tbl:dropout}
\resizebox{0.60\columnwidth}{!}{
{\normalsize
\setlength{\arrayrulewidth}{.06em}
\begin{tabular}{@{}c|c|ccc}
Dropout &  Acc & \multicolumn{3} {c}{F1@\{25,50,75\}} \\ \hline
0.0 & 92.0 & 80.8    &  71.9    &  52.8 \\
0.1 & 91.7 & 83.0   & 73.4  & 53.8     \\
0.2 & 91.6 & 84.3   & 74.7  & 54.6     \\
0.3 & \B 92.1 &  \B 84.9      & \B 75.7    & \B 56.0        \\
0.4 & 91.6 &  84.7      & 75.2    & 54.9       \\ 
0.5 & 91.5 &  83.6      & 73.6    & 53.5       \\

\end{tabular}
}
}
\end{table}

\subsection{Comparison to Previous Work}
In this experiment, we compare our model with the method introduced in \cite{momeni2020watch}, which uses a cosine function as the similarity measurement between the \emph{query} and the \emph{target} features. If the similarity value is above a predefined threshold, the \emph{query} will be regarded as spotted. Their provided pretrained model is trained jointly on BSL-1K \cite{albanie2020bsl} and BSLDict datasets. However, neither the BSL-1K dataset nor the training code are accessible publicly. Hence, we apply their pretrained model on the BSLCORPUS using the common dictionary between BSLDict and SignBank, while also noting the possibility of getting improved results if trained on the applied datasets. The threshold used in this experiment is 0.7. The results are shown in Table~\ref{tbl:bslcorp}. From the table, we see that our method significantly outperforms Watch-Read-Lookup \cite{momeni2020watch} on the sign spotting task.

\begin{table}[!h]\centering
\ra{1.05}
\caption{Comparison against the state-of-the-art on the BSLCORPUS dataset.}\label{tbl:bslcorp}
\resizebox{0.88\columnwidth}{!}{
{\normalsize
\setlength{\arrayrulewidth}{.06em}
\begin{tabular}{@{}l|c|ccc}
Methods & Acc & \multicolumn{3} {c}{F1@\{25,50,75\}}      \\ \hline
Watch-Read-Lookup \cite{momeni2020watch} & 78.6 & 46.5   & 31.1  & 10.9     \\
Our method & \B 92.1 &  \B 85.6      & \B 76.4    & \B 56.3  \\
\end{tabular}
}
}
\end{table}

\subsection{Transfer to Other Sign Language}
To evaluate the generalization of our model to different datasets, we apply \SignLookup~to the German Sign Language dataset: PHOENIX14T. This dataset has frame-level annotation for glosses, however, it lacks the corresponding sign dictionary for use as the \emph{query}. To overcome this, we use one instance of each gloss category from the training subset as the \emph{query}. As there might be multiple instances for one category, we choose the example with a frame length closest to 16. In total, there are 1198 different isolated signs extracted as the dictionary.

\begin{table}[!h]\centering
\ra{1.05}
\caption{Results of transferring our model to the Pheonix2014T dataset.}\label{tbl:pheonix2014T}
\resizebox{0.85\columnwidth}{!}{
{\normalsize
\setlength{\arrayrulewidth}{.06em}
\begin{tabular}{@{}l|c|ccc}
Dataset  & Acc & \multicolumn{3} {c}{F1@\{25,50,75\}}    \\ \hline

 Pheonix2014T - DEV & 96.1 & 93.3   & 89.8  & 76.8     \\
 Pheonix2014T - TEST & 95.6 & 93.6 & 89.1 & 75.7 
\end{tabular}
}
}
\end{table}

The performance of our model applied to the PHOENIX14T dataset is shown in Table ~\ref{tbl:pheonix2014T}. From the table, we can see that it outperforms the BSLCORPUS dataset tests. This is because we are using continuous videos from the same corpus for both \emph{query} and \emph{target} and this means that the spatial and temporal domain differences are minimal. However, these numbers are potentially worse that they should be as the gloss annotation on PHOENIX14T (as published) is automatically generated and therefore incorporates annotation noise.

\section{Qualitative Experiments}
\label{sect:qual}
In this section, we report qualitative results for the \SignLookup~model. We share snapshot examples of spotting a sign in Figure~\ref{fig:example}, with more examples provided in supplementary material. On the left hand side, is shown the  isolated sign video we wish to use as the \emph{query}. On the right side, it is the continuous \emph{target} video we wish to identify the sign in. In the \emph{target} video, the top bar (labeled gt) shows the ground truth with red depicting the start and stop of the sign in the larger continuous sequence. The bottom bar shows our prediction in green. The curve denotes the probability that the queried sign is spotted. In the upper-left corner of both videos, the corresponding gloss annotation is shown. 

In the first example, we see that our model can accurately identify the \emph{query} in the \emph{target} video even though they come from two different datasets that have large spatial and temporal domain differences. In the second example, we apply our model to a DGS Pheonix2014T example, demonstrating generalization to sign languages and dataset.

\begin{figure}[t]
\begin{center}
  \includegraphics[width=0.8\linewidth]{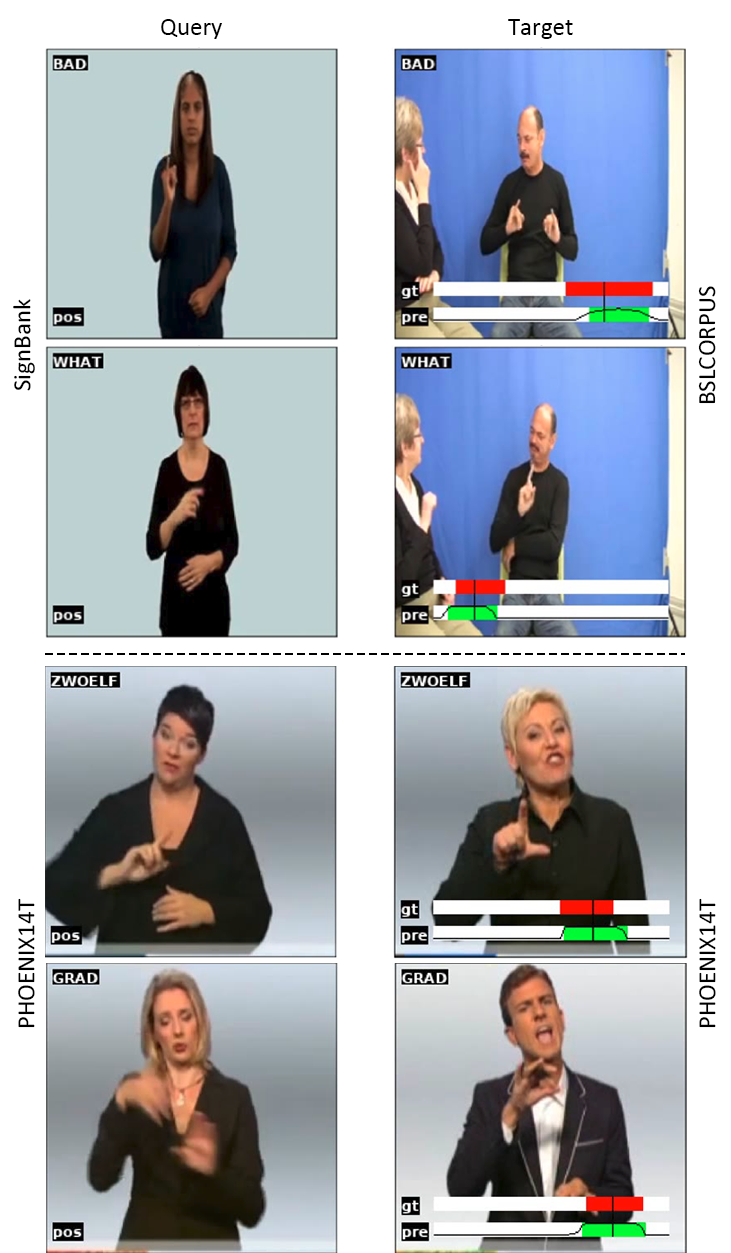} 
\end{center}
  \caption{Some qualitative sign spotting examples produced by our model. The results in the top two rows use the SignBank dataset for the \emph{query} (left) and BSLCORPUS dataset for the \emph{target} (right). The results in the bottom two row uses PHOENIX14T for both the \emph{query} and \emph{target}. }
\label{fig:example}
\end{figure}

\section{Conclusion}
\label{sect:conc}
In this paper, we proposed a novel transformer-based network called \SignLookup, which given a single isolated example of a sign, can identify or 'spot' occurrences of that sign in continuous real world signing footage. This not only solves the temporal domain difference between the \emph{query} and the \emph{target} videos, but also addresses the alignment between the sliding windows and sign boundaries. The experiments showed that \SignLookup~improves the accuracy of sign spotting, especially when we have temporal scale discrepancy, achieving state-of-the-art performance. By also applying \SignLookup~to a popular DGS dataset, we demonstrate its capability to generalise to other sign languages. 

In the future, we would like to expand our approach to multi-channel streams including hands, body pose and facial expression and mouthings. We also plan to release this as a tool for linguists to help with annotation.

\section{Acknowledgments}
This work received funding from the SNSF Sinergia project ‘SMILE II’ (CRSII5 193686), the European
Union’s Horizon2020 research and innovation programme
under grant agreement no. 101016982 ‘EASIER’ and the
EPSRC project ‘ExTOL’ (EP/R03298X/1). This work reflects only the authors view and the Commission is not responsible for any use that may be made of the information
it contains.


{\small
\bibliographystyle{ieee}
\bibliography{egbib}
}

\end{document}